# RMSE-ELM: Recursive Model based Selective Ensemble of Extreme Learning Machines for Robustness Improvement

Bo Han[a], Bo He[a,*], Mengmeng Ma[a], Tingting Sun[a],

Tianhong Yan[b,*], Amaury Lendasse [c,d]

[a]*School of Information and Engineering, Ocean University of China, Shandong, Qingdao, China 266000*
[b]*School of mechanical and Electrical Engineering, China Jiliang University, Zhejiang, Hangzhou, China 310018*
[c]*Department of Mechanical and Industrial Engineering and the Iowa Informatics Initiative, The University of Iowa, Iowa City, IA 52242-1527, USA*
[d]*Arcada University of Applied Sciences, 00550 Helsinki, Finland*

___________________________________________________________________

**Abstract**

Extreme Learning Machine (ELM) as an emerging branch of shallow networks has shown its excellent generalization and fast learning speed. However, for blended data, the robustness of ELM is so weak because its weights and biases of hidden nodes are set randomly. Moreover, the noisy data exert a negative effect. To solve this problem, a new framework called "RMSE-ELM" is proposed in this paper. It is a two-layer recursive model. In the first layer, the framework trains lots of ELMs in different ensemble groups concurrently, then employs selective ensemble approach to pick out an optimal set of ELMs in each group, which can be merged into a large group of ELMs called candidate pool. In the second layer, selective ensemble approach is recursively used on candidate pool to acquire the final ensemble. In the experiments, we apply UCI blended datasets to confirm the robustness of our new approach in two key aspects (Mean Square Error and Standard Deviation). The space complexity of our method is increased to some degree, but the results have shown that RMSE-ELM significantly improves robustness with a rapid learning speed compared to representative methods (ELM, OP-ELM, GASEN-ELM, GASEN-BP and E-GASEN). It becomes a potential framework to solve robustness issue of ELM for high-dimensional blended data in the future.

*Key words:* Extreme Learning Machine, Recursive Model, Selective Ensemble, RMSE-ELM, Robustness Improvement

___________________________________________________________________
_______

\* Corresponding author
   *Email address:* bhe@ouc.edu.cn (Bo He); thyan@163.com (Tianhong Yan)

# 1 Introduction

In recent two or three decades, neural networks are increasingly popular in machine learning community. Especially for recent five years, lots of researchers mainly have paid their attention on deep structures such as Deep Boltzmann Machine [1], Convolution Neural Network [2] and so on. However, the deep networks are hardly applied into real-time area in big data era because of two reasons: First of all, there is no free lunch in any algorithms. Though the training accuracy of deep network is pretty high, the training time is so long that we can hardly bear the computational cost [3]. Secondly, the deep structures tend to fall into the pit called "over-fitting", which means that it has a bad generalization. What's more, the tuning of parameters in deep networks is very time-consuming [4]. So the shallow structure is naturally our intuition for big data analysis and real-time application.

Recently, the Extreme Learning Machine (ELM) [5] as an emerging branch of shallow networks was proposed by Guang-Bin Huang et. al. It was evolved from single hidden layer feed-forward networks (SFLNs). It has shown the excellent generalization performance and fast learning speed compared to Deep Belief Networks [6] or Deep Boltzmann Machines [7]. In essence, the algorithm of ELM has two main steps: In the first step, the input weights and biases can be assigned randomly, which will definitely reduce computational cost because they do not need to be tuned manually. In the second step, the output weights of ELM can be computed easily by the generalized inverse of hidden layer output matrix and target matrix [8]. In terms of the computational performance of ELM, it tends to reach not only the smallest training error but also the smallest norm of output weights with rapid speed. Based on above merits of ELM, a lot of researchers in machine learning community now increasingly customize their own frameworks based on ELM for specific issues. For equalization problems, ELM based complex-valued neural networks are a powerful tool. For regression or multi-label issues, the kernel based ELM proposed by Huang et. al is effective [9,10]. For generalization problem, Incremental ELM [11] outperforms many representative algorithms like SVM [12], stochastic BP [13] and so on. What's more, various extended ELMs also attract our attention. For example, online sequential ELM [14] is an efficient learning algorithm to handle both additive [15] and RBF [16,17] nodes in the unified framework. In complex dimensional space, the kernel implementation of ELM is superior to conventional SVM. From the above discussion, we can conclude that ELM is an excellent algorithm for different issues in machine learning area.

However, as the keynote given by Guang-Bin Huang indicates, the robustness analysis is still one of the open problem in ELM community [5,18]. Different researchers have different research styles to tackle with the same problem. Previously, Rong et.al presented pruning algorithm called P-ELM to improve the robustness of ELM [19]. And also Miche and Lendasse, proposed an algorithm called OP-ELM [20,21] to improve the robustness due to its variable selection mechanism, which removes the irrelevant variables from blended data efficiently [22,23]. However, for blended data (namely the raw data is blended with noisy data), they do not work very well because of two reasons. First, the mechanism of variables pruning is very time-consuming. What's more, the standard deviations of training error in above two models are relatively high, which means that these models are not the top choice for robustness improvement. If we want to improve the robustness of original ELM, we should initially clarify why the ELM is so weak for blended data. First of all, we believe ELM sets its initial weights and biases randomly, which largely reduce the computational time but cannot guarantee the suitable parameters of hidden nodes for good robustness. Second, the noisy data exert a negative effect on robustness of ELM. So for blended data, my initial intuition is that if we train a batch of different ELMs and then ensemble them averagely, we might improve the robustness because of Hansen and Salamon's theory [24]. It proved that the robustness performance of a single network can be improved by an ensemble of neural networks. Sollich and Krogh [25] confirmed it later. Thus, based on this theory, Sun et. al proposed the average weighted ELM ensemble [26], which has a better generalization than original ELM on raw data. But on blended data, the average weighted ELM ensemble does not work well because it is negatively affected by noisy data such as Gaussian noise or Uniform noise. Zhou et. al [27] proposed a new framework called GASEN, which can resist the negative effect from noisy data. In his theory, the ensemble of several optimal networks may be better than the ensemble of all networks. The GASEN is fully based on genetic algorithm and Back-Propagation (BP) neural networks. Therefore, in real-time area, we should not apply GASEN directly for robustness improvement because of high computation cost.

Inspired by above observations, for blended data [28], we hope to create a new computational framework, which not only improves the robustness largely but also keeps a rapid learning speed. So in this paper, a new approach called "RMSE-ELM" is proposed. Our tuition can be concluded into two aspects: First, selective ensemble approach is an effective tool to resist noisy data but the kernel of framework is usually the BP networks. What's more, the genetic algorithm itself is a little bit complicated. Therefore, the training process is so time-consuming [29]. So we hope to employ the advantage of ELM to speed up the selective ensemble approach. Second, in cognitive science, the information processing of human brain is constructed hierarchically, and it can extract different useful information layer by layer. However, the more layers we construct, the more parameters the algorithm will learn, which will definitely increase the computational cost. Therefore, we hope to construct a semi-shallow framework for a good compromise between robustness and computational cost. For technical details, it is a two-layer recursive model. In the first layer, we concurrently train lots of ELMs in different groups, then we employ selective ensemble approach to pick out several ELMs in each group, which can be transmitted into the second layer called candidates pool. In the second layer, we employ selective ensemble approach recursively to pick out several ELMs for the average ensemble. In the experiments, we apply UCI blended datasets [30] to confirm the robustness of new method, which is compared to that of several methods such as ELM, OP-ELM, GASEN-ELM, GASEN-BP and E-GASEN in two key aspects: Mean Square Error and Standard Deviation. Though the space complexity of our method is increased to some degree, the results have shown that the RMSE-ELM significantly improves the robustness with a rapid learning speed. We will further explore how many layers can achieve the optimal compromise between the robustness and computational cost in our framework. The extended RMSE-ELM has a great potential to be a trend framework to solve robustness issue of ELM for high-dimensional blended data in the future.

We organize the rest of the paper as follows. In Section 2, we discuss previous work on classical ELM and Selective Ensemble. In Section 3 we describe our new method called RMSE-ELM from structure to theory. In Section 4, for UCI blended datasets, several experimental results on ELM, OP-ELM, GASEN-ELM, GASEN-BP, E-GASEN are reported respectively. In Section 5, we present our discussions the motivation of benchmark selection and other facts revealed by experiments. Finally, in Section 6, conclusions are drawn and future work and direction are indicated.

## 2 Previous Works

### 2.1 Extreme Learning Machine

Extreme learning machine (ELM) has been developed to obtain a much faster learning speed and higher generalization performance both in the regression and classification problem. The essence of ELM is the hidden layers of SFLNs need not to be tuned iteratively [5,31], that is, the parameters of the hidden nodes which include input weights and biases can be randomly generated, and then it only needs to solve the output weights. The structure of ELM is shown below.

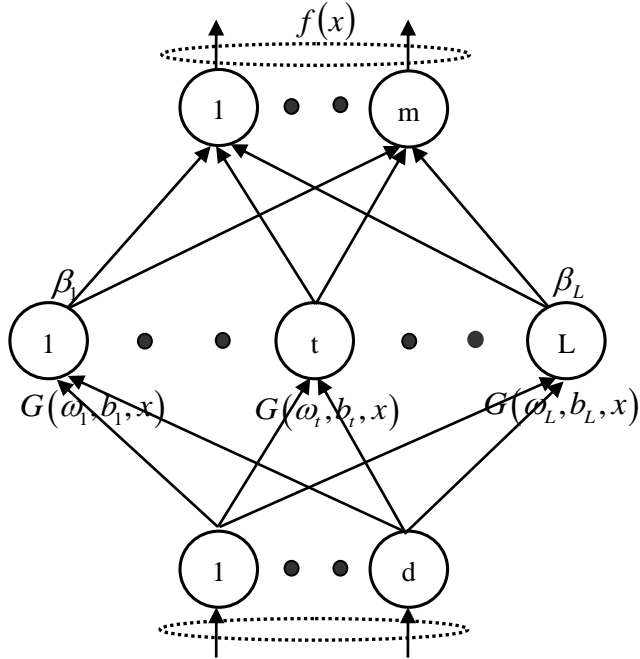

Figure 1. The structure of ELM algorithm

For the given $N$ learning samples $\{x_i, y_i\}_{i=1}^{N}$, where $x_i = [x_{i1}, ..., x_{id}]'$ and $y_i = [y_{i1}, ..., y_{im}]'$ the standard model of the ELM learning with $L$ hidden neurons and activation function $G(\omega_t, b_t, x_i)$ can be written as

$$\sum_{t=1}^{L} \beta_t\, G(\omega_t, b_t, x_i) = o_i, \quad i = 1, \cdots, N \tag{1}$$

Where $\omega_t = [\omega_{t1}, \ldots, \omega_{td}]'$ is the weight vector connecting the $t\_th$ hidden neuron and the input neurons. $\beta_t = [\beta_{t1}, \ldots, \beta_{tm}]'$ denotes the weight vector connecting the $t\_th$ hidden neuron and the output neurons. $b_j$ is the bias of the $t\_th$ hidden neuron. ELM can approximate these $N$ samples with zero error means that

$$\sum_{i=1}^{N} \|o_i - y_i\| = 0 \tag{2}$$

Namely, there exist $(\omega_j, b_j)$ and $\beta_j$ such that

$$\sum_{t=1}^{L} \beta_t\, G(\omega_t, b_t, x_i) = y_i, \quad i = 1, \cdots, N \tag{3}$$

The activation function $G(\omega_t, b_t, x_i)$ can be arbitrarily chosen from the sigmoid function, the Hard-limit function, the Gaussian function, the Multi-quadric function and any other function which is infinitely differentiable in any interval so that the hidden layer parameters can be randomly generated. The above equation can also be written compactly as:

$$H\beta = Y \tag{4}$$

Where

$$H = \begin{bmatrix} G(\omega_1, b_1, x_1) & \cdots & G(\omega_L, b_L, x_1) \\ & \vdots & \\ G(\omega_1, b_1, x_N) & \cdots & G(\omega_L, b_L, x_N) \end{bmatrix}_{N \times L} \tag{5}$$

$$\beta = [\beta'_1, \ldots, \beta'_L]'_{L \times m} \tag{6}$$

$$Y = [y'_1, \ldots, y'_N]'_{N \times m} \tag{7}$$

Here $H$ is called the hidden layer output matrix of the neural network. When the training set $x_i$ is given and the parameters $(\omega_t, b_t)$ are randomly generated, matrix $H$ can be obtained. And then the output weights $\beta$ can be generated as:

$$\beta = H^\dagger Y \tag{8}$$

Where $H^\dagger$ denotes the Moore-Penrose generalized inverse of matrix $H$ [32,33].

In summary, the ELM algorithm can be presented as follows:

**Algorithm 1 Extreme Learning Machine**

**Input:** The $N$ training set $\{x_i, y_i\}_{i=1}^{N}$ , the activation function $G(\omega_t, b_t, x_i)$, and the number of hidden nodes $L$.

**Steps:**

1. Randomly generate input weights $\omega_t$ and biases $b_t, t = 1, ..., L$
2. Calculate the hidden layer output matrix $H$.
3. Calculate the output weight vector $\beta = H^\dagger Y$.

### 2.2 Selective Ensemble

In recent years, ensemble learning has received lots of attention from machine learning community due to its potential to improve the generalization capability of a learning system [34,35]. With the increase of size, the prediction speed of an ensemble machine decreases significantly but its storage increases quickly. Z.H Zhou et. al[36] has proved that many could be better than all and proposed a new framework called selective ensemble. The aim of selective ensemble learning is to further improve the prediction accuracy of an ensemble machine, to enhance its prediction speed as well as to decrease its storage need. Selective ensemble learning mainly involves three steps [37]:

(1) Training a set of base learners individually generated from bootstrap samples of a fixed training data.

(2) Selecting right components from all the available learners and excluding the bad base learners to form an optimal ensemble. Genetic algorithm is used for components selection. The population of base learners is encoded as real chromosomes so that one bit represents the average weight of initial learner ensemble. Suppose $x$ is randomly sampled through a distribution $p(x)$, and the expected output is $y$, and the output of the $ith$ base ELM is $f_i(x)$. The optimum weight $\omega^*$ is expressed as empirical Eq(9) which minimizes the generalization error of the ensemble model.

$$\omega^* = arg\min_{\omega} \left( \sum_{i=1}^{N} \sum_{j=1}^{N} \omega_i \omega_j C_{ij} \right) \quad (9)$$

$C_{ij}$ is the correlation between the $ith$ and the $jth$ individual base learner. And the definition is as follows.

$$C_{ij} = \int dx p(x) (f_i(x) - y)(f_j(x) - y) \quad (10)$$

Therefore, the $kth$ $(k = 1 ..., N)$ of optimum weight $\omega^*$ can be solved by Lagrange multiplier, which satisfies Eq(11):

$$\omega_k^* = \frac{\sum_{j=1}^{N} c_{kj}^{-1}}{\sum_{i=1}^{N} \sum_{j=1}^{N} c_{ij}^{-1}} \quad (11)$$

Genetic algorithm based selective ensemble assigns a random weight to every base ELM first. Then, genetic algorithm is used to evolve those weights so that they can characterize the fitness of the ELM in joining the ensemble to some extent.

(3) Combining the selected base learner components to get the final predictions.

### 3 New Method

## 3.1 The Structure of RMSE-ELM

Inspired by above discussions, for blended data, we hope to create a new computational framework, which not only improves the robustness performance of ELM largely but also keeps a rapid learning speed. We naturally have two tuitions below.

First of all, Traditional selective ensemble approach like GASEN algorithm is definitely an effective tool to resist noisy data because it utilizes fewer but better individual models to ensemble, which achieves stronger generalization ability. But both genetic algorithm employed by GASEN and the training process of individual kernels (BPs) are so time-consuming, which can hardly be used in industry or real-time situation. So we hope to build our customized selective ensemble based on ELM kernels because of its rapid learning speed.

Secondly, from the point view of cognitive science, the information processing of human brain is constructed hierarchically, and it can extract different useful information layer by layer. However, if we completely construct our networks as our brain, for example a deep-layer network, we may encounter several training problems. Firstly, the training time is so long that we can rarely bear the computational cost, not to mention big data analysis. Secondly, the deep structures tend to fall into the pit called "over-fitting" which in turn means the weak generalization. Moreover, the tuning of parameters in deep networks needs large amount of time and personal experience. So the semi-shallow structure is naturally top choice for big data analysis and real-time application.

In this paper, we present a framework called "RMSE-ELM" to improve the robustness of ELM for blended data with acceptable computational cost. The figure of our framework shows in below.

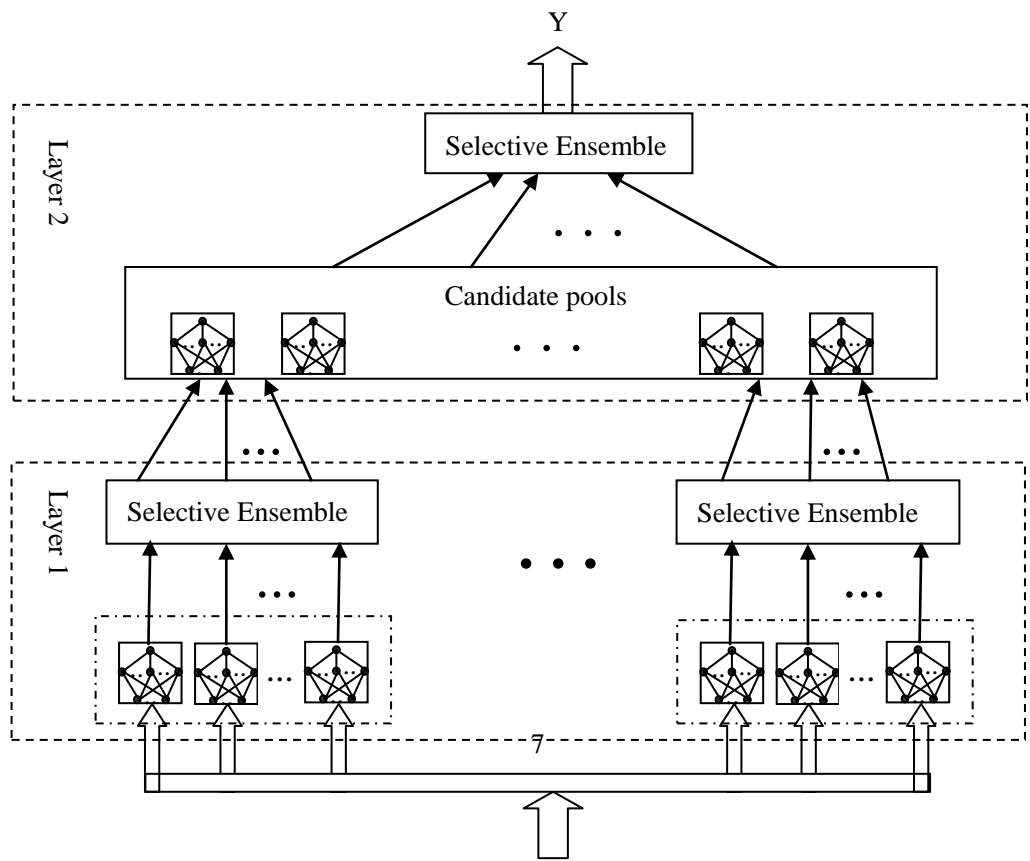

Figure 2. The Framework of RMSE-ELM

Just as the Figure 2, it is a two-layer recursive model, which is a good compromise between shallow and deep network. In the first layer, we concurrently train lots of ELMs that belong to the different ensemble groups, then we employ selective ensemble approach to pick out several ELMs in each group, which can be transmitted into our second layer – the pool of better candidates. In the second layer, we employ selective ensemble recursively to pick from selected ELMs and then ensemble an optimal set of ELMs to acquire the final result.

Although our framework is relatively simple compared with deep structure networks, we believe that it locates in the right track to solve the robustness issues of ELM.

**3.2 The Theory of RMSE-ELM**

Now let's first analyze our framework in theory. From above discussion, we can clearly see our framework recursively employ selective ensemble approach. In essence, the recursive model algorithm based selective ensemble can be explained as the hierarchical model based selective ensemble. So if the selective ensemble can work well, theoretically, the recursive model based selective ensemble can work better.

So firstly we should analyze whether the selective ensemble of extreme learning machine are good enough. Please note currently the individual networks are ELMs instead of BP networks. To be honest, it is not an easy task excluding the bad ELMs from our target group. In order to generate the ensemble ELM with small size but stronger generation ability, genetic algorithm is used to select the ELM models with high fitness from a set of available ELMs. Suppose that the learning task is to approximate a function $f: R^m \to R^n$, it can be represented by an ensemble of $N$ base ELM learners. The predictions of the base ELM learners are combined by weighted averaging, where a weight $\omega_i (i = 1 \dots N)$ is assigned to the individual base ELM learner $f_i (i = 1 \dots N)$, and $\omega_i$ satisfies Eq(12).

$$0 \leq \omega_i \leq 1, \quad \sum_{i=1}^{N} \omega_i = 1 \tag{12}$$

Then the output of ensemble is:

$$\bar{f}(x) = \sum_{i=1}^{N} \omega_i f_i(x) \tag{13}$$

Where $f_i$ is the output of the $ith$ base ELM learner.

We assume that each base ELM learner has only one output. Suppose $x \in R^m$ is randomly sampled through a distribution $p(x)$. And the target for $x$ is $d(x)$. Then the error $E_i(x)$ of the $ith$ base ELM learner and the error $E(x)$ of the ensemble on input $x$ are respectively:

$$E_i(x) = \big(f_i(x) - d(x)\big)^2 \tag{14}$$

$$E(x) = \big(\bar{f}(x) - d(x)\big)^2 \tag{15}$$

Then the generalization error $E_i$ of the $ith$ base ELM learner and the generalization error $E$ of the ensemble on the distribution $p(x)$ are respectively:

$$E_i = \int dx p(x) E_i(x) \tag{16}$$

$$E = \int dx p(x) E(x) \tag{17}$$

Define the correlation between the $ith$ and the $jth$ individual base ELM learner as:

$$C_{ij} = \int dx p(x) \big(f_i(x) - d(x)\big)\big(f_j(x) - d(x)\big) \tag{18}$$

Apparently, $C_{ij}$ satisfies:

$$C_{ii} = E_i \text{ and } C_{ij} = C_{ji} \tag{19}$$

According to Eq(13) and Eq(15):

$$E(x) = \left(\sum_{i=1}^{N} \omega_i f_i(x) - d(x)\right)\left(\sum_{j=1}^{N} \omega_j f_j(x) - d(x)\right) \tag{20}$$

Then according Eq(17), Eq(18) and Eq(20):

$$E = \sum_{i=1}^{N}\sum_{j=1}^{N} \omega_i \omega_j C_{ij} \tag{21}$$

When the base ELM learners are combined by the simple ensemble method, that is $\omega_i = \frac{1}{N}$ for every $i$, we have

$$E = \sum_{i=1}^{N}\sum_{j=1}^{N} C_{ij}/N^2 \tag{22}$$

Now, we assume that the $kth$ base learner is omitted, the new generalization error $\widehat{E}$:

$$\widehat{E} = \sum_{\substack{i=1 \\ i \neq k}}^{N}\sum_{\substack{j=1 \\ j \neq k}}^{N} C_{ij}/(N-1)^2 \tag{23}$$

According to Eq(16), the generalization error of the $kth$ base ELM learner:

$$E_k = \int dx p(x) E_k(x) \tag{24}$$

Therefore,

$$E - \widehat{E} = \frac{2\sum_{\substack{i=1 \\ i \neq k}}^{N} C_{ik} + E_k - (2N-1)E}{(N-1)^2} \tag{25}$$

So if

$$2\sum_{\substack{i=1 \\ i \neq k}}^{N} C_{ij} + E_k - (2N-1)E > 0 \tag{26}$$

Then,

$$E > \hat{E} \tag{27}$$

Which means new ensemble omitting the $kth$ learner is now more robust than original ensemble.

So we can get a constraint condition from Eq(26) and Eq(27),

$$(2N-1)\hat{E} < (2N-1)E < 2\sum_{\substack{i=1\\i\neq k}}^{N} C_{ik} + E_k \tag{28}$$

If we multiply Eq(28) by $(N-1)^2$,

$$(2N-1)(N-1)^2\hat{E} < 2(N-1)^2\sum_{\substack{i=1\\i\neq k}}^{N} C_{ik} + (N-1)^2 E_k \tag{29}$$

According to Eq(23) and Eq(29), the constraint condition can be deduced as follows.

$$(2N-1)\sum_{\substack{i=1\\i\neq k}}^{N}\sum_{\substack{j=1\\j\neq k}}^{N} C_{ij} < 2(N-1)^2\sum_{\substack{i=1\\i\neq k}}^{N} C_{ik} + (N-1)^2 E_k \tag{30}$$

Therefore, it is proved that when using the simple ensemble method and when constraint condition Eq(30) is satisfied, then omitting the $kth$ base learner will improve the ensemble's generalization ability.

There is a conclusion that after lots of ELMs are trained, ensembling an appropriate subset of them is superior to ensembling all of them in some cases. The individual ELMs that should be omitted satisfy Eq(30). This result implies that the ensemble does not use all the networks to achieve good performance. Therefore, the selective ensemble of ELM can work well.

According to the above proofs, the recursive model based selective ensemble of extreme learning machine might be better than the selective ensemble of extreme learning machine because of three reasons below: firstly, the best result comes from the better results more easily, so if the first layer of our framework can effectively select an optimal group of different ELMs, the second layer has a great potential to produce a better result based on an optimal group of ELMs. Secondly, from the network structure, the recursive model based selective ensemble can be explained as the hierarchical model based selective ensemble. And the RMSE-ELM is a natural extension of selective ensemble of extreme learning machine. Therefore, if each part can work well, the whole system can work well at least. Finally, lots of experiments in recent years have shown that if more neural networks are included, in some cases the generalization error of the ensemble might be further reduced.

From above theoretical discussion, we see that why the recursive model based selective ensemble of extreme learning machine can work better. However, we will further explore how many layers can achieve the optimal compromise between robustness and computational cost. The pseudo code of our current framework is organized as follows:

**Algorithm 2 RMSE-ELM**

**Given:** training set $(X, Y)$, $M$(the size of ensemble groups in the first layer), $N_1$(the size of each ensemble in the first layer), $N_2$(the size of candidates pool in the second layer), $\omega^*$ is defined in Eq(9), threshold $\lambda$ is a pre-set value (reciprocal value of $N_1$ or $N_2$).

**Steps:**

1. for $group = 1 ... M$
   { $N_2 = 0$ ;
      for $element = 1 ... N_1$
      { Training each ELM network;
       Generating a population of weight vector;
       Using selective ensample to get the best weight vector $\omega_1^*$;
       Removing base ELMs that the weights less than $\lambda_1 = 1/N_1$;
      }
      Calculating the whole remained ELMs of group $i$ are $n_i$;
   $$N_2 = N_2 + n_i;$$
   }

2. Training $N_2$ remained ELM;
3. Using selective ensemble to get the best weight vector $\omega_2^*$;
4. Removing base ELMs that the weights less than $\lambda_2 = 1/N_2$;
5. Getting the final prediction;

## 4 Experiments

In this section, we present some experiments on 4 UCI blended datasets to verify whether RMSE-ELM performs better in robustness than other methods such as ELM, OP-ELM, GASEN-ELM, GASEN-BP and E-GASEN for blended data. At the same time, computational cost is also a significant parameter to evaluate the usefulness of our new framework. All simulations are carried out in Matlab environment running in an Intel Corei5-3470 (3.20GHz CPU).

Table 1. Specification of the 2 tested regression data sets

| Task | # variables | # training | # test | Abbr. |
|---|---|---|---|---|
| Boston Housing | 13 | 400 | 106 | BH |
| Abalone | 8 | 2000 | 2177 | Aba |
| Red Wine | 11 | 1065 | 534 | RW |
| Waveform | 21 | 3000 | 2000 | Wav |

Four types of datasets are all selected from the UCI machine learning repository [39]. The first one is Boston Housing dataset which contains 506 samples. Each sample is composed of 13 input variables and 1 output variable. And this dataset is divided into a training set of 400 samples and a testing set of the rest. The second one is Abalone dataset. There are 7 continuous input variables, 1 discrete input variable and 1 categorical attribute in this dataset. It comprises 4177 samples, among which, 2000 samples are used for training and the rest 2177 samples are used for testing. The third one is Red Wine dataset which contains 1599 samples. Each sample consists of 11 input variables and 1 output variable, the dataset is divided into two sections: 1065 samples for training set and the rest samples for testing set. Finally, Waveform dataset with more number of input variables is selected. This dataset contains 21 input variables and 1 output variable. The specification of the four types of datasets is shown in table 1.

Firstly, we randomly mix several irrelevant Gaussian noises with the original UCI data, and all features of data are normalized into a similar scale. Secondly, we train the different models such as ELM, OP-ELM, GASEN-ELM, GASEN-BP, E-GASEN and RMSE-ELM on the training set of blended data. Finally, we test the different models on the testing set of blended data to acquire experimental results including Mean Square Error (MSE), Standard Deviation (STD) and Computational Cost (CC). In our experiments, the genetic algorithm employed by RMSE-ELM is implemented by the GAOT toolbox developed by Houck et al. In the toolbox, the genetic operators (selecting, crossover probability, mutation probability and stopping criterion) are set to the default values. The first group of original UCI data is blended with 7 irrelevant variables that all conform to the Gaussian distributions, such as $N(0,2)$, $N(0,1)$, $N(0,0.5)$, $N(0,0.1)$, $N(0,0.005)$, $N(0,0.001)$, $N(0,0.0005)$. To acquire the convincing result, the second group of original data is blended with 10 irrelevant Gaussian variables, such as $N(0,2)$, $N(0,1)$, $N(0,0.5)$, $N(0,0.1)$, $N(0,0.05)$, $N(0,0.01)$, $N(0,0.005)$, $N(0,0.001)$, $N(0,0.0005)$, $N(0,0.0001)$. For different ensemble frameworks (GASEN-ELM, GASEN-BP, E-GASEN and RMSE-ELM), The number of ELMs in each ensemble group is initially set to 20 [38], so the threshold $\lambda$ used by selective ensemble is set to 0.05 because it is the reciprocal value of the size of each ensemble according to Zhou's experiment. For hierarchical models such as E-GASEN and RMSE-ELM, the number of ensemble groups is set to 4 according to the Zhou's experiments. In addition, the number of hidden units in each ELM is set to 50 because it can acquire the better performance at this point. Specifically speaking, the testing RMSE curve gradually decreases to a constant value and also the learning time is still less after this point [40]. For each algorithm we perform 5 runs and record the average value of MSE, STD and CC. The experimental results are shown in following tables and figures.

Table 2. MSE for UCI blended datasets(7 irrelevant variables)

| Data set | ELM | OP-ELM | GASEN-ELM | GASEN-BP | E-GASEN | **RMSE-ELM** |
|---|---|---|---|---|---|---|
| BH | 5.8564 | 4.9823 | 5.0543 | 4.7869 | 4.8822 | **4.7763** |
| Aba | 34.5586 | 31.4742 | 30.0193 | 29.5716 | 28.3969 | **26.0626** |
| RW | 0.4998 | 0.4946 | 0.4514 | 0.5412 | 0.4488 | **0.4374** |
| Wav | 0.3733 | 0.3412 | 0.3429 | 0.2671 | 0.3371 | **0.3276** |

Table 3. MSE for UCI blended datasets (10 irrelevant variables)

| Data set | ELM | OP-ELM | GASEN-ELM | GASEN-BP | E-GASEN | **RMSE-ELM** |
|---|---|---|---|---|---|---|
| BH | 6.3748 | 5.0672 | 5.7973 | 4.8495 | 5.6263 | **5.4462** |
| Aba | 34.7401 | 29.5260 | 29.7477 | 27.6825 | 27.5196 | **26.2389** |
| RW | 0.5069 | 0.4969 | 0.4613 | 0.5399 | 0.4512 | **0.4422** |
| Wav | 0.3750 | 0.3339 | 0.3489 | 0.2747 | 0.3449 | **0.3347** |

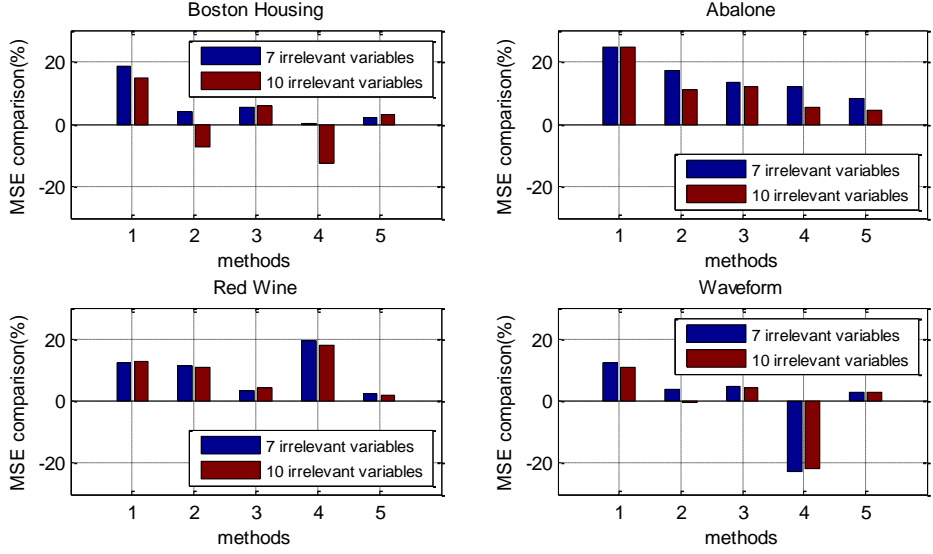

Figure 3. MSE comparison between RMSE-ELM and other methods (x-axis 1:ELM, 2:OP-ELM, 3:GASEN-ELM, 4:GASEN-BP, 5:E-GASEN)

There are two important criteria for robustness assessment (MSE and STD). Let's first analyze the MSE among different methods on UCI blended datasets. For the evaluation of MSE, we visualize the experimental results in Table 2 and Table 3 into Figure 3. We define the difference of MSE between RMSE-ELM and other methods as MSE comparison. The formula is

$$\text{MSE comparison} = \frac{\text{MSE(other methods)} - \text{MSE(RMSE\_ELM)}}{\text{MSE(other methods)}} \times 100\% \quad (31)$$

Therefore, in Figure 3, positive percentage means the MSE of new method (RMSE-ELM) is lower than other methods, which in turn proves that the robustness of new method is better, or vice versa. In four types of UCI blended datasets, the results show that the MSE of our method is lower than that of other methods in most cases. In particular, the difference of MSE between our method and ELM is more obvious, which definitely proves that our framework improves the robustness performance of original ELM for blended data. However, in some cases, the MSE of GASEN-BP and OP-ELM is obviously lower than that of RMSE-ELM.

Table 4. STD for UCI blended datasets (7 irrelevant variables)

| Data set | ELM | OP-ELM | GASEN-ELM | GASEN-BP | E-GASEN | **RMSE-ELM** |
|---|---|---|---|---|---|---|
| BH | 0.2236 | 0.1416 | 0.1024 | 0.1551 | 0.0494 | **0.1109** |
| Aba | 3.2644 | 7.2611 | 1.3031 | 1.6831 | 0.4601 | **1.3439** |
| RW | 0.0191 | 0.0091 | 0.0092 | 0.0270 | 0.0033 | **0.0110** |
| Wav | 0.0094 | 0.0187 | 0.0031 | 0.0069 | 0.0020 | **0.0041** |

Table 5. STD for UCI blended datasets (10 irrelevant variables)

| Data set | ELM | OP-ELM | GASEN-ELM | GASEN-BP | E-GASEN | **RMSE-ELM** |
|---|---|---|---|---|---|---|
| BH | 0.1864 | 0.1807 | 0.0923 | 0.1702 | 0.0400 | **0.1047** |
| Aba | 3.1029 | 4.3826 | 1.7374 | 1.8569 | 0.4019 | **1.4385** |
| RW | 0.0168 | 0.0166 | 0.0086 | 0.0216 | 0.0023 | **0.0085** |
| Wav | 0.0107 | 0.0233 | 0.0039 | 0.0098 | 0.0016 | **0.0026** |

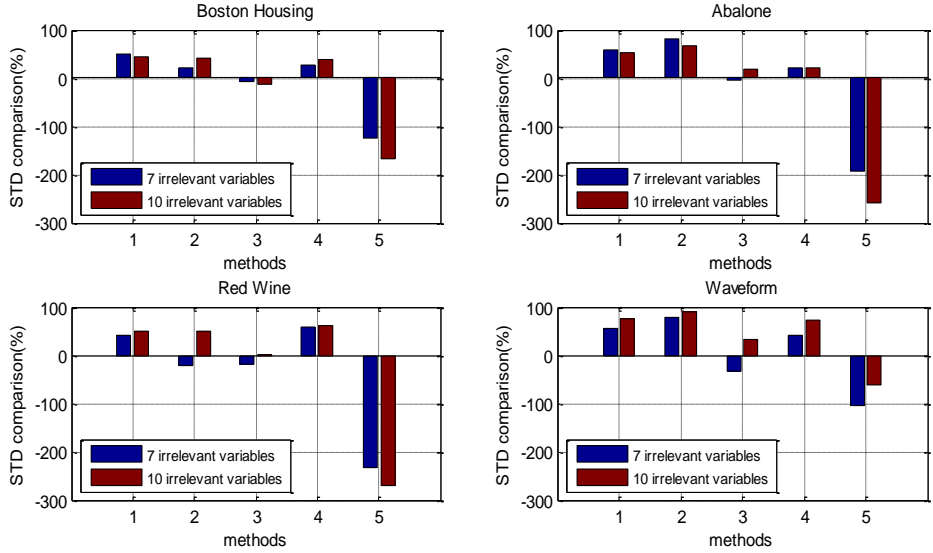

Figure 4. STD comparison between RMSE-ELM and other methods (x-axis 1:ELM, 2:OP-ELM, 3:GASEN-ELM, 4:GASEN-BP, 5:E-GASEN)

Secondly, for the evaluation of STD, we visualize the experimental results in Table 4 and Table 5 into Figure 4. We define the difference of STD between RMSE-ELM and other methods as STD comparison. The formula is

$$\text{STD comparison} = \frac{\text{STD(other methods)} - \text{STD(RMSE\_ELM)}}{\text{STD(other methods)}} \times 100\% \quad (32)$$

In Figure 4, positive percentage means the STD of our method is lower than that of other methods, which proves that the robustness of our new method is better, or vice versa. In four types of blended datasets, the results show that the STD of our method is lower than that of other methods, which confirms that our framework really improve the robustness performance for blended data. However, in some cases, the STD of E-GASEN is obviously lower than that in RMSE-ELM.

Table 6. CC for UCI blended datasets (7 irrelevant variables, unit: seconds)

| Data set | ELM | OP-ELM | GASEN-ELM | GASEN-BP | E-GASEN | **RMSE-ELM** |
|---|---|---|---|---|---|---|
| BH | 0.0920 | 234.5413 | 2.5023 | 574.1617 | 4.6832 | **3.7206** |
| Aba | 0.0250 | 25.7682 | 1.4180 | 205.4845 | 7.6893 | **2.4960** |
| RW | 0.0390 | 189.7191 | 1.8720 | 361.7819 | 3.0015 | **2.9203** |
| Wav | 0.1427 | 534.6310 | 2.8408 | 1534.0000 | 4.8984 | **3.8485** |

Table 7. CC for UCI blended datasets (10 irrelevant variables, unit: seconds)

| Data set | ELM | OP-ELM | GASEN-ELM | GASEN-BP | E-GASEN | **RMSE-ELM** |
|---|---|---|---|---|---|---|
| BH | 0.0952 | 281.5818 | 2.7363 | 634.8929 | 3.8517 | **3.9226** |
| Aba | 0.0250 | 33.0161 | 1.4383 | 229.8675 | 6.8874 | **2.7191** |
| RW | 0.0406 | 263.2673 | 1.7581 | 431.6392 | 2.3665 | **3.0373** |
| Wav | 0.1045 | 559.4664 | 2.7924 | 1995.4000 | 6.2244 | **3.8454** |

Finally, according to Table 6 and Table 7, the results show that the CC of our method is acceptable. However, the CC of GASEN-BP and OP-ELM is too long to apply in the real-time area or industry.

There are two interesting observations above, and we hope to explain further. Firstly, although in some cases, the MSE of GASEN-BP and OP-ELM is lower than that of RMSE-ELM, from the view of statistics, the MSE of RMSE-ELM is lower than that of GASEN-BP and OP-ELM on the whole. For example, we have 4 types of UCI datasets and 2 types of Gaussian noisy variants. If we run above 3 algorithms on 8 types of blended data, for MSE comparison between RMSE-ELM and GASEN-BP, the MSE of RMSE-ELM is lower on 5 types of blended data while the MSE of GASEN-BP is lower on 3 types of blended data. For MSE comparison between RMSE-ELM and OP-ELM, the MSE of RMSE-ELM is lower on 6 types of blended data while the OP-ELM is lower on only 2 types of blended data. What's more, the CC of RMSE-ELM is much shorter than that of OP-ELM and GASEN-BP. Secondly, in some cases, though the STD of E-GASEN is lower than that of RMSE-ELM, the MSE of RMSE-ELM is totally lower than that of E-GASEN. Moreover, the CC of RMSE-ELM is shorter than that of E-GASEN except RW dataset for 10 irrelevant noisy variables.

In conclusion, we believe that our new method in robustness is definitely better than ELM. We believe that our framework is a good compromise between robustness performance and learning speed. However, how many groups in the first layer of RMSE-ELM should we choose for the best robustness performances? It should be further explored.

**5 Discussions**

Until now, we are very clear about the structure and performance of RMSE-ELM. In the design of experiments, for added noises, the Gaussian noises are selected because they are common in real world. For comparable methods, we select OP-ELM as one of the benchmark methods because it is almost the first generation of extended ELM to probe the robustness issue. And both the GASEN-ELM and E-GASEN are also selected because they have the similar mechanism with RMSE-ELM. However, the differences in structure and mechanism among them are also obvious. For example, GASEN-ELM is a one-layer ensemble network using selective ensemble approach.

Though the E-GASEN is a two-layer ensemble network like RMSE-ELM, the ensemble in the second layer is regarded as the simple ensemble instead of the selective ensemble approach employed by RMSE-ELM. According to the selection of UCI blended data and benchmark approaches, we believe that our experimental results should be fair and convincing.

In the experiments, we tested new method on four types of UCI datasets, which are blended with 7-dimensional and 10-dimensional Gaussian noises separately. It is clear that the MSE of our method is almost lower than that of other methods except for GASEN-BP in some cases. For GASEN-BP and RMSE-ELM, the CC of GASEN-BP limit its wide use in industry and real-time area compared with RMSE-ELM. And also the STD of our method is lower than that of other methods except for E-GASEN. For E-GASEN and RMSE-ELM, though the E-GASEN is lower in STD, which means that E-GASEN is more stable in fluctuation of MSE, in the rest aspects (MSE and CC), the performance of E-GASEN is totally worse than that of RMSE-ELM. In conclusion, the robustness performance of our method is than that of other methods for blended data with relatively fast speed. In essence, the ELM has a weak robustness performance for blended data mainly because of its simple structure, so the hierarchical model like recursive model inference is our natural consideration.

## 6 Conclusions

In this paper, we proposed a new method called RMSE-ELM. To be more specific, the structure of our framework is the two-layer ensemble architecture, which recursively employs selective ensemble to pick out several optimal ELMs from bottom to top for the final ensemble. The experiments prove that the robustness performance of RMSE-ELM is better than original ELM and representative methods for blended data. Through analysis of experiments, the reasons why our approach works are proposed as follows. Firstly, the selective ensemble extracts the optimal subset effectively from each group in the first layer and from candidate pool in the second layer. Secondly, the kernel of our framework is ELM, which has excellent generalization and rapid learning speed. Finally, the recursive model in essence is a special case of hierarchical network, which is a good compromise between shallow network and deep network. However, analyses presented in this paper are very preliminary. More experiments and principles still need to be completed in order to modify our framework further. Our future work will focus on three main directions: First, in the framework of RMSE-ELM, how many groups in the first layer should we choose to acquire the best robustness. And how many layers can achieve the optimal compromise between robustness and computational cost based on our framework. Second, whether the space complexity of our method can be largely reduced under regularized framework. For example, if the weight of our framework can be sparse enough under regularization, the complexity of our framework might be largely reduced. Third, whether the selective ensemble approach in the top layer can be replaced by other criteria for a better robustness performance. In general, it may be an interesting work to develop a combination of ensemble learning and hierarchical model to enhance the robustness performance of ELM in the future.

**Acknowledgments**

This work is partially supported by Natural Science Foundation of China (41176076, 31202036, 51075377).

**References**

[1]  R. Salakhutdinov, G.-E. Hinton, Deep BoltzmannMachine, 12th International Conference on Artificial Intelligence and Statistics Proceedings (AISTATS), 5 (2009) 448-455.

[2]  C.-C. Han, C.-T.Wang, B.-S.Jenget. al, The Application of a Convolution Neural Network on Face and License Plate Detection, 12th International Conference on Pattern Recognition, 3 (2006) 552-555.

[3] G.-E. Hinton, S. Osindero, Y.-W.Teh, A fast learning algorithm for deep belief nets, Neural Computation, 18(2006) 1527-1554.

[4]  Y. Bengio, Learning deep architectures for AI, Foundations and trends in machine learning,2(2009) 1-127.

[5]  G.-B. Huang, Q.-Y. Zhu, C.-K. Siew, Extreme learning machine: theory and applications, Neurocomputing,70 (2006) 489-501.

[6]  G.-E. Hinton, R. Salakhutdinov, Reducing the dimensionality of Data eith Niural Networks, Science, 313 (2006) 504-507

[7]  R. Salakhutdinov, H. Larochelle, Efficient learning of deep boltzmann machines, International Conference on Artificial Intelligence and Statistics, (2010).

[8]  G.-B. Huang, D.-H. Wang, Y. Lan, Extreme learning machines: a survey, International Journal of Machine Learning and Cybernetics,2 (2011) 107-122.

[9]  G.-B. Huang, C.-K. Siew, Extreme learning machine with randomly assigned RBF kernels, Eighth International Conference on Control, Automation, Robotics and Vision Proceedings (ICARCV), (2004).

[10] B. Fr énay, M. Verleysen, Parameter-insensitive kernel in extreme learning for non-linear support vector regression, Neurocomputing, 74 (2011) 2526-2531.

[11] G.-B. Huang, L. Chen, C.-K. Siew, Universal Approximation using incremental constructive feedforward networks with random hidden node, IEEE Transactions on Neural Networks, 17 (2006) 879-892.

[12]C. Schuldt, I. Laptev, B. Caputo, Recognizing human axtions: a local SVM approach, 17th International Conference on Pattern Recognitionproceedings, 3 (2004) 32-36.

[13] D.-E. Rumelhart, G.-E. Hinton, R.-J. Williams, Learning representations by back-propagation errors, Nature, 323 (1986) 533-536.

[14] N.-Y. Liang, G.-B. Huang, P. Saratchandranet al., A fast and accurate on-line

sequential learning algorithmfor feedforward networks,IEEE Transactions on Neural Networks, 17 (2006) 1411-1423.

[15] L.-C. Yu, L. Bottou, G.-B. Orr et. al, Efficient backprop, Lect Notes Computer Science, 1524 (1998) 9-50.

[16] G.-B. Huang, P. Saratchandran, N. Sundararajan, An efficient sequential learning algorithm for growing and pruning RBF (GAPP-RBF) networks,IEEE Transitions Man Cybernet, 34 (2004) 2284-2292.

[17] G.-B. Huang, P. Saratchandran, N. Sundararajan, A generalized growing and pruning RBF (GGAP-RBF) neural network for function approximation, IEEE Transitions Neural Network, 16 (2005) 57-67.

[18] G.-B. Huang, D.-H. Wang, Y. Lan, Extreme learning machines: a survey, International Journal of Machine Learning and Cybernetics,2 (2011) 107-122.

[19] H.-J. Rong, Y.-S. Ong, A.-H. Tan et. al, A fast pruned-extreme learning machine for classification problem, Neurocomputing, 72 (2008) 359-366.

[20] Y. Miche, A. Sorjamaa, P. Bas, et. al, OP-ELM: optimally pruned extreme learning machine, IEEE Transactions on Neural Networks, 21 (2010) 158-162.

[21] Y. Miche, A. Sorjamaa, A. Lendasse, OP-ELM: theory, experiments and a toolbox,Artificial Neural Networks-ICANN 2008, ed: Springer, 2008, pp. 145-154.

[22] Y. Miche, A. Sorjamaa, A. Lendasse, OP-ELM: theory, experiments and a toolbox,Artificial Neural Networks-ICANN 2008, ed: Springer, 2008, pp. 145-154.

[23] Y. Miche, P. Bas, C. Jutten, et. al, A methodology for building regression models using extreme learning machine: OP-ELM, Proceedings of the European Symposium on Artificial Neural Networks (ESANN), (2008) 247-252.

[24] L. K. Hansen, P. Salamon, Neural network ensembles, IEEE Transactions on Pattern Analysis and Machine Intelligence, 12 (1990) 993-1001.

[25] A. Krogh, P. Sollich, Statistical mechanics of ensemble learning, The American Physical Society, (1997).

[26] Z.-L. Sun, T.-M. Choi, K.-F. Au, et. al, Sales forecasting using extreme learning machine with applications in fashion retailing, Decision Support Systems, 46 (2008) 411-419.

[27] Z.-H. Zhou. J.-X. Wu, J. Yuan, et. al, Genetic algorithm based selective neural network ensemble, IJCAI-01: Proceedings of the Seventeenth International Joint Conference on Artificial Intelligence, Seattle, Washington, August 4-10, 2001, pp. 797.

[28] Y. Tang, B. Biondi, Least-squares migration/inversion of blended data, SEG Technical Program Expanded Abstracts, 2009, pp. 2859-2863.

[29] N. Li, Z.-H. Zhou, Selective ensemble under regularization framework Multiple

Classifier Systems, ed: Springer, 2009, pp. 293-303.

[30] A. Asuncion, D.- J. Newman, UCI machine learning repository, 2007.

[31] G.-B. Huang, Q.-Y. Zhu, C.-K. Siew, Extreme learning machine: a new learning scheme of feedforward neural networks, IEEE International Joint Conference onNeural Networks, 2 (2004) 985-990.

[32] D. Serre, Matrices: Theory and Applications. 2002, ed: Springer, New York, 2002.

[33] C.-R. Rao, S.-K. Mitra, Generalized inverse of a matrix and its applications, J. Wiley, New York, 1972.

[34] H.-M. Van, Y. Miche, E. Oja et. al, Adaptive ensemble models of extreme learning machines for time series prediction, Lecture Notes Computing Science, 5769 (2009) 305-314.

[35] H.-M. Van, Y. Miche, E. Oja et. al, Gpuaccelerated and parallelized ELM ensembles for large-scale regression, Neurocomputing, (2011).

[36] Z.-H. Zhou, J.-X. Wu, W. Tang, Ensemble neural networks: Many could be better than one, Artificial Intelligence, 137(2002) 239-263.

[37] L.-J. Zhao, T.-Y. Chai, D.-C. Yuan, Selective ensemble extreme learning machine modeling of effluent quality in wastewater treatment plants, International Journal of Automation and Computing, 9(2012) 627-633.

[38] Opitz D W, Shavlik J W, Generating accurate and diverse members of a neural-network ensemble, Advances in neural information processing systems, (1996) 535-541.

[39] Asuncion, Arthur, and David Newman, UCI machine learning repository, (2007).

[40] Huang, Guang-Bin, Lei Chen, and Chee-Kheong Siew. Universal approximation using incremental constructive feedforward networks with random hidden nodes. Neural Networks, IEEE Transactions on 17.4 (2006): 879-892.